# Convolutional Recurrent Neural Networks for Small-Footprint Keyword Spotting

*Sercan Ö. Arık[1,\*], Markus Kliegl[1,\*], Rewon Child[1], Joel Hestness[1], Andrew Gibiansky[1], Chris Fougner[1], Ryan Prenger[1], Adam Coates[1]*

[1]Baidu Silicon Valley Artificial Intelligence Lab, 1195 Bordeaux Dr. Sunnyvale, CA 94089, USA
[\*]Equal contribution

sercanarik@baidu.com, klieglmarkus@baidu.com

## Abstract

Keyword spotting (KWS) constitutes a major component of human-technology interfaces. Maximizing the detection accuracy at a low false alarm (FA) rate, while minimizing the footprint size, latency and complexity are the goals for KWS. Towards achieving them, we study Convolutional Recurrent Neural Networks (CRNNs). Inspired by large-scale state-of-the-art speech recognition systems, we combine the strengths of convolutional layers and recurrent layers to exploit local structure and long-range context. We analyze the effect of architecture parameters, and propose training strategies to improve performance. With only ~230k parameters, our CRNN model yields acceptably low latency, and achieves 97.71% accuracy at 0.5 FA/hour for 5 dB signal-to-noise ratio.

**Index Terms**: Keyword spotting, speech recognition, convolutional neural networks, recurrent neural networks.

## 1. Introduction

Motivated by the most common way humans interact with each other, conversational human-technology interfaces are becoming increasingly popular in numerous applications. High-performance speech-to-text conversion and text-to-speech conversion constitute two important aspects of such interfaces, as most computational algorithms are developed for text inputs and outputs. Another crucial aspect of conversational interfaces is keyword spotting (KWS) – also known as wakeword detection, to enable transitioning between different computational states based on the voice input provided by the users. KWS systems aim to detect a particular keyword from a continuous stream of audio. As their output determines different states of the device, very high detection accuracy for a very low false alarm (FA) rate is critical to enable satisfactory user experience. Typical applications exist in environments with interference from background audio, reverberation distortion, and the sounds generated by the speaker of the device in which the KWS is embedded. A KWS system should demonstrate robust performance in this wide range of situations. Furthermore, the computational complexity and model size are important concerns for KWS systems, as they are typically embedded in consumer devices with limited memory and computational resources, such as smartphones or smart-home sensors.

There are already millions of devices with embedded KWS systems. Traditional approaches for KWS are based on Hidden Markov Models with sequence search algorithms [1]. With the advances in deep learning and increase in the amount of available data, state-of-the-art KWS has been replaced by deep learning-based approaches due to their superior performance [2]. Deep learning-based KWS systems commonly use Deep Neural Networks (DNNs) combined with compression techniques [3,4] or multi-style training approaches [5,6]. A potential drawback of DNNs is that they ignore the structure and context of the input, and an audio input can have strong dependencies in time or frequency domains. With the goal of exploiting such local connectivity patterns by shared weights, Convolutional Neural Networks (CNNs) were explored for KWS [7,8]. A potential drawback of CNNs is that they cannot model the context over the entire frame without wide filters or great depth. Recurrent neural networks (RNNs) were also studied for KWS with connectionist temporal classification (CTC) loss [9,10], unlike the aforementioned DNN and CNN models [2-6] with cross-entropy (CE) loss. Yet, a high accuracy at a low FA rate could not be obtained, given the ambitious targets of the applications of such systems. Similar to DNNs, a potential limitation of RNNs is that the modeling is done on the input features, without learning the structure between successive time and frequency steps. Recently, [11] proposed a Convolutional Recurrent Neural Network (CRNN) architecture with CTC loss. However, despite the large model size, similar to RNNs, a high accuracy at a low FA rate could not be obtained.

In this paper, we focus on developing a production-quality KWS system using CRNNs with CE loss for a small-footprint model, applied for a single keyword. Our goal is to combine the strengths of CNNs and RNNs, with additional strategies applied during training to improve the overall performance, while keeping a small-footprint size. The rest of the paper is organized as follows. In Section 2, we describe the end-to-end architecture and training methodologies for small-footprint KWS. In Section 3, we explain the experiments and the corresponding results. In Section 4, we present our conclusions.

## 2. Small-footprint keyword spotting

### 2.1. End-to-end architecture

We focus on a canonical CRNN architecture, inspired by the successful large-scale speech recognition systems [12-14]. To adapt these architectures for small-footprint KWS, the model size needs to be shrunk two to three orders of magnitude. We will analyze the impact of different parameters on performance while shrinking the size of the model.

Fig. 1 shows the CRNN architecture with the corresponding parameters. The raw time-domain inputs are converted to per-channel energy normalized (PCEN) mel spectrograms [8], for succinct representation and efficient training. (Other input representations we experimented with yielded worse performance for model architectures of comparable size.) The 2-D PCEN features are given as inputs to the convolutional layer, which employ 2-D filtering along both the time and frequency dimensions. The outputs of the convolutional layer are fed to bidirectional recurrent layers, which might include gated recurrent units (GRUs) [15] or long short-term memory (LSTM) units [16] and process the entire frame. Outputs of the recurrent layers are given to the fully connected (FC) layer. Lastly, softmax decoding is applied over two neurons, to obtain a corresponding scalar score. We use rectified linear units as activation function in all layers.

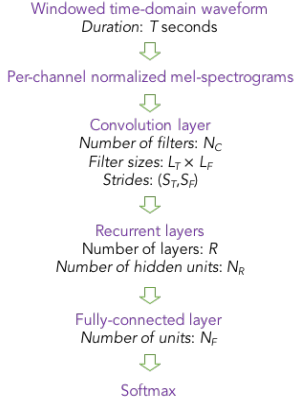

Figure 1: *End-to-end CRNN architecture for KWS.*

### 2.2. End-to-end training

**Algorithm 1** Sequential alignment of keyword samples
**require:** keyword characters $c_k$ ($1 \leq k \leq K$), smoothed character occupancy scores $p(c_k, t)$, decay rate $\alpha$ ($0 \leq \alpha \leq 1$)
**initialize:** $p^{lr}(c_k, t) = p(c_k, t)$, $p^{rl}(c_k, t) = p(c_k, t)$
**for:** $n := 1$ to $N_{iter}$
 **for:** $k := 0$ to $K-2$ (right-to-left decoding)
  $T^{rl}_{c_k} = \arg\max_t p^{rl}(c_k, t)$
  $p^{rl}(c_{k+1}, t) = \alpha \cdot p^{rl}(c_{k+1}, t)$ for $t \geq T^{rl}_{c_k}$
 **end**
 **for:** $k := K-1$ to 1 (left-to-right decoding)
  $T^{lr}_{c_k} = \arg\max_t p^{lr}(c_k, t)$
  $p^{lr}(c_{k-1}, t) = \alpha \cdot p^{lr}(c_{k-1}, t)$ for $t \leq T^{lr}_{c_k}$
 **end**
**end**
**return**: $(\min(T^{lr}_{c_1}, T^{rl}_{c_1}), \max(T^{lr}_{c_K}, T^{rl}_{c_K}))$

In speech recognition, large-scale architectures with recurrent layers typically use variants of CTC loss to decode the most probable output label. Aside from the modeling limitations due to conditional independence assumptions of targets, CTC loss has a high computational complexity and typically yields good performance only when the model capacity is sufficiently large to efficiently learn from a large dataset. As we focus on small-footprint architectures, the loss function that is optimized during the training is chosen as the CE loss for the estimated and target binary labels, indicating whether a frame corresponds to a keyword or not.

To train with a CE loss, unlike CTC, precise alignment of the training samples is important. We use Deep Speech 2 [14], a large-scale speech recognition model, to obtain the estimated probability distributions of keyword characters $c_k$ ($1 \leq k \leq K$) for each time instance. As the CTC decoding yields peaked distributions, we further smooth the output over time and obtain smoothed character occupancy scores $p(c_k, t)$. We then obtain the beginning and end times of the keywords using the heuristic algorithm shown in Algorithm 1. An extra short padding is added while chopping the keywords to cover edge cases. The accuracy of alignments obtained were significantly beyond the time scale of human perception.

## 3. Experiments and Results

### 3.1. Data and training

We develop our KWS system for the keyword "TalkType" (which can be pronounced as a single word or two words). We choose a frame length of $T = 1.5$ seconds, which is sufficiently long to capture a reasonable pronunciation of "TalkType". Using a sampling rate of 16 kHz, each frame contains 24k raw time-domain samples. Corresponding PCEN mel spectrograms are obtained for 10 ms stride and 40 channels, yielding an input dimensionality of 40 × 151. The entire data set consists of ~16k different samples, collected from more than 5k speakers. The dataset is split into training, development and test sets with 6-1-1 ratio. Training samples are augmented by applying additive noise, with a power determined by a signal-to-noise ratio (SNR) sampled from [-5,15] dB interval. The additive noise is sampled from a data set of representative background noise and speech, with a total length exceeding 300 hours. To provide robustness against alignment errors, training samples are also augmented by introducing random timing jitter. We use the ADAM optimization algorithm for training [17], with a batch size of 64. The learning rate is initially chosen as 0.001, and later dropped to 0.0003.

Our evaluation considers a streaming scenario such that inference is performed for overlapping frames of duration $T$. The shift between the frames is chosen as 100 ms, (which should ideally be much longer than the spectrogram stride and much shorter than the inference latency - see Section 3.2 for more details). The metrics we focus on are the false rejection rate (FRR) and false alarms (FA) per hour, typically fixing the latter at a desired value such as 1 FA/hr [7]. Noise is added to the development and test sets, with a magnitude depending on the SNR value. We note that the collected samples are already noisy so the actual SNR is lower if defined precisely as the ratio of powers of the information-bearing signal and the noise. Similar to our augmentation of the training sets, negative samples and noise datasets are sampled from representative background noise and speech.

### 3.2. Impact of the model architecture

Table 1 shows the performance of various CRNN architectures for the development set with 5 dB SNR. We note that all models were trained until convergence, even though it requires very different number of epochs. We observe the general trend that the larger model size typically yields better performance. Increasing the number of convolution filters or increasing the number of recurrent hidden units are the two effective approaches to improve the performance. Increasing the number of recurrent layers has a limited impact, and GRU is preferred over LSTM as a better performance can be obtained for a lower complexity.

Table 1: *Performance of different CRNN architectures (see Fig. 1 for the description of the parameters). The chosen set of parameters for the rest of the paper is colored and highlighted in bold.*

| Convolutional | | | Recurrent | | | FC | Total number of parameters | FRR (%) for the noise development set with 5 dB SNR | |
|---|---|---|---|---|---|---|---|---|---|
| $N_C$ | $(L_T, L_F)$ | $(S_T, S_F)$ | $R$ | $N_R$ | Recurrent unit | $N_F$ | | at 1 FA/hour | at 0.5 FA/hour |
| 32 | (20,5) | (8,2) | 2 | 8 | GRU | 32 | 45k | 5.54 | 7.44 |
| 32 | (20,5) | (8,2) | 3 | 8 | LSTM | 64 | 68k | 6.17 | 7.68 |
| 32 | (5,1) | (4,1) | 2 | 8 | GRU | 64 | 102k | 6.04 | 7.31 |
| 32 | (20,5) | (8,2) | 2 | 16 | GRU | 64 | 110k | 3.48 | 4.46 |
| 32 | (20,5) | (20,5) | 2 | 32 | GRU | 64 | 110k | 5.70 | 7.99 |
| 32 | (20,5) | (8,2) | 3 | 16 | GRU | 64 | 115k | 3.42 | 4.10 |
| 16 | (20,5) | (8,2) | 2 | 32 | GRU | 32 | 127k | 3.53 | 5.55 |
| 32 | (20,5) | (12,4) | 2 | 32 | GRU | 64 | 143k | 5.80 | 7.72 |
| 16 | (20,5) | (8,2) | 1 | 32 | GRU | 64 | 148k | 4.20 | 6.27 |
| 128 | (20,5) | (8,2) | 3 | 8 | GRU | 32 | 159k | 3.83 | 5.21 |
| 64 | (10,3) | (8,2) | 1 | 16 | GRU | 32 | 166k | 3.21 | 4.31 |
| 128 | (20,5) | (8,2) | 1 | 32 | LSTM | 64 | 197k | 3.37 | 4.56 |
| 32 | (20,5) | (12,2) | 2 | 32 | GRU | 64 | 205k | 3.26 | 4.40 |
| 32 | (20,5) | (8,2) | 1 | 32 | GRU | 64 | 211k | 3.00 | 3.84 |
| **32** | **(20,5)** | **(8,2)** | **2** | **32** | **GRU** | **64** | **229k** | **2.85** | **3.79** |
| 32 | (40,10) | (8,2) | 2 | 32 | GRU | 64 | 239k | 3.57 | 5.03 |
| 32 | (20,5) | (8,2) | 3 | 32 | GRU | 64 | 248k | 3.00 | 3.42 |
| 32 | (20,5) | (8,2) | 2 | 32 | LSTM | 64 | 279k | 3.06 | 4.41 |
| 32 | (20,5) | (8,1) | 2 | 32 | GRU | 64 | 352k | 2.23 | 3.31 |
| 64 | (20,5) | (8,2) | 2 | 32 | GRU | 64 | 355k | 2.43 | 3.99 |
| 64 | (20,5) | (8,2) | 2 | 32 | LSTM | 32 | 407k | 3.11 | 4.04 |
| 64 | (10,3) | (4,1) | 2 | 32 | GRU | 64 | 674k | 3.37 | 4.35 |
| 128 | (20,5) | (8,2) | 2 | 32 | GRU | 128 | 686k | 2.64 | 3.78 |
| 32 | (20,5) | (8,2) | 2 | 128 | GRU | 128 | 1513k | 2.23 | 2.95 |
| 256 | (20,5) | (8,2) | 4 | 64 | GRU | 128 | 2551k | 2.18 | 3.42 |
| 128 | (20,5) | (4,1) | 4 | 64 | GRU | 128 | 2850k | 2.64 | 3.21 |

It is desired to limit the model size given the resource constraints for inference latency, memory, and power consumption. Following [7], we choose the size limit as 250k (which is more than 6 times smaller than the architecture with CTC loss in [11]). For the rest of the paper, the default architecture is the set of parameters highlighted in bold, which also corresponds to a fairly optimal point given the model size vs. performance trade-off.

We compare the performance with a CNN architecture based on [7]. Given the discrepancy in input dimensionality and training data, we reoptimize the model hyperparameters for the best performance while upper-bounding the number of parameters to 250k for a fair comparison. For the same development set with 5 dB SNR, the best CNN architecture achieves 4.31% FRR at 1 FA/hour and 5.73% FRR at 0.5 FA/hour. Both metrics are ~51% higher compared to the FRR values of the chosen CRNN model with 229k parameters. Interestingly, the performance gap is lower for higher SNR values. We elaborate on this in Section 3.4.

Recall that the model is bidirectional and runs on overlapping 1.5 second windows at 100 ms stride. However, thanks to the small model size and the large time stride of 8 in the initial convolution layer, we are able to do inference comfortably faster than real time. The inference computational complexity of the chosen CRNN-based KWS model with 229k parameters is roughly ~30M floating point operations (FLOPs) when implemented on processors of modern consumer devices (without special functions to implement nonlinear operations). Even when implemented on modern smartphones without any approximations and special function units, our KWS model can achieve an inference time much faster than the time scale for reactive time for humans with auditory stimuli, which is ~280 ms [18].

### 3.3. Impact of the amount of training data

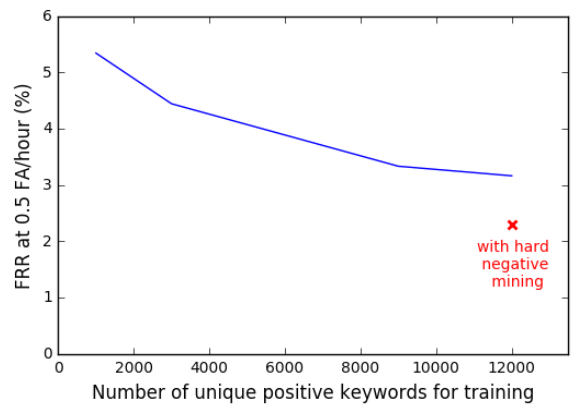

Figure 2: *FRR at 0.5 FA/hour vs. number of unique training keywords for the test set with 5 dB SNR.*

Given the representation capacity limit imposed by the architecture size, increasing the amount of positive samples in the training data has a limited effect on the performance. Fig. 2 shows the FRR at 0.5 FA/hour (for the test set with 5 dB SNR) vs. the number of unique "TalkType" samples used while training. Saturation of performance occurs faster than applications with similar type of data but with large-scale models, e.g. [14].

Besides increasing the amount of the positive samples, we observe performance improvement by increasing the diversity of relevant negative samples, obtained by hard mining. We mine negative samples, by using the pre-converged model on a very large public videos dataset (that are not used in training, development, or test sets). Then, training is continued using the mined negative samples until convergence. As shown in Fig. 2, hard negative mining yields decrease in FRR for the test set.

### 3.4. Noise robustness

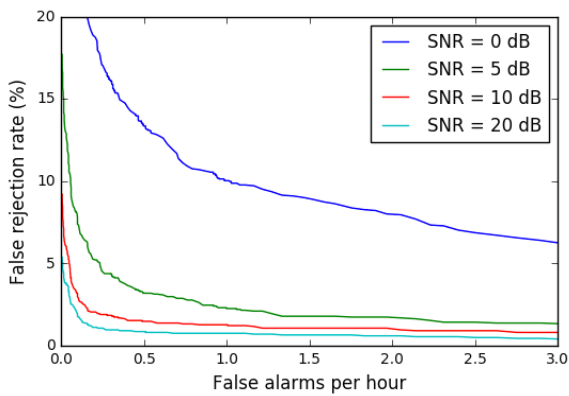

Figure 3: *FRR vs. FA per hour for the test set with various SNR values.*

For the test set with various SNR values, Fig. 3 shows the FRR vs. FA per hour. For higher SNR, lower FRR is obtained, and stable performance starts for a lower FA rate. Note that the SNR values (in dB) of the augmented training samples are sampled from a distribution with a mean of 5 dB, and deterioration in performance is observed beyond this value. Performance for lower SNR values can be improved by augmenting with lower SNR, but this comes at the expense of decreased performance for higher SNR, which can be attributed to the limited learning capacity of the model.

We observe the benefit of recurrent layers especially for lower SNR values. The performance gap of CRNN architectures with CNN architectures (adapted from [7] as explained in Section 3.1) reduces as the SNR increases. We hypothesize that the recurrent layers are better able to adapt to the noise signature of individual samples, since each layer processes information from the entire frame. CNNs, in contrast, require wide filters and/or great depth for this level of information propagation.

### 3.5. Far-field robustness

Our dataset already consists of samples recorded at varying distance values, which should be representative for most applications such as smartphone KWS systems. Yet, some applications, such as smart-home KWS systems, require high performance at far-field conditions.

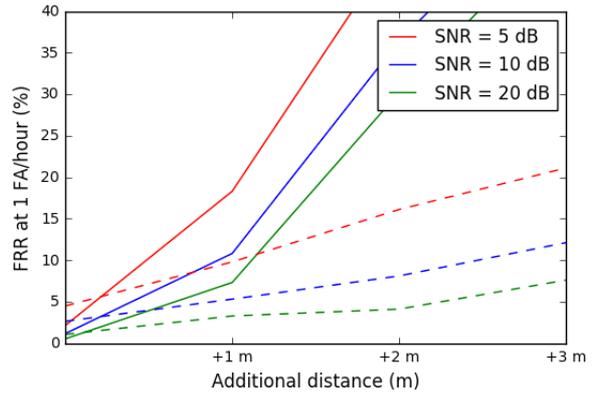

Figure 4: *FRR at 1 FA/hour vs. additional distance for far-field test sets with varying SNR values. Solid: baseline performance, dashed: with far-field augmented training.*

Fig. 4 shows performance degradation with the additional distance. Far-field test sets are constructed by augmenting the original test set with impulse responses corresponding to a variety of configurations at the given distance (considering different values for degrees of arrival etc.). Significant deterioration in performance is observed especially in conjunction with higher noise, as also explained in [19]. To provide robustness against this deterioration, we consider training with far-field-augmented training samples, using a variety of impulse responses that are different than the ones in the test set. This augmentation achieves significantly less degradation in performance for farther distances. Yet, it yields a worse performance for the original data set due to the training/testing mismatch.

## 4. Conclusions

We studied CRNNs for small-footprint KWS systems. We presented the trade-off between model size and performance, and demonstrated the optimal choice of parameters given the tradeoff. The capacity limitation of the model has various implications. Performance gain is limited by merely increasing the number of positive samples, yet hard negative mining improves the performance. Training sets should be carefully chosen to reflect the application environment, such as the noise level or far-field conditions. Overall, at 0.5 FA/hour (which is an acceptable value from a user perspective), our model achieves 97.71%, 98.71% and 99.3% accuracy for the test set with 5 dB, 10 dB and 20 dB SNR values, respectively. Our numerical performance results may seem better than other KWS models in the literature. However, a direct comparison is not meaningful because of the difference in the datasets and the actual keywords, i.e. the inference task. Given that human performance is excellent in the KWS task, we still believe that there is further room for improvement in terms of performance.

## 5. Acknowledgements

Discussions with Andrew Ng, Sanjeev Satheesh, Jiaji Huang, Jue Sun, and Bing Jiang are gratefully acknowledged. We thank Hui Song for the impulse response measurements used for far-field augmentation.

# 6. References

[1] J.R. Rohlicek, W. Russell, S. Roukos, and H. Gish, "Continuous hidden Markov modeling for speaker-independent wordspotting," in *IEEE Proceedings of the International Conference on Acoustics, Speech and Signal Processing*, 1990, pp. 627–630.

[2] G. Chen, C. Parada, and G. Heigold, "Small-footprint keyword spotting using deep neural networks," in *Proceedings International Conference on Acoustics, Speech, and Signal Processing*, 2014, pp. 4087-4091.

[3] G. Tucker, M. Wu, M. Sun, S. Panchapagesan, G. Fu, and S. Vitaladevuni, "Model compression applied to small-footprint keyword spotting," in *Proceedings of Interspeech*, 2016, pp. 1393-1397

[4] Vikas Sindhwani, Tara N. Sainath, and Sanjiv Kumar, "Structured transforms for small-footprint deep learning," in *Neural Information Processing Systems*, 2015, pp. 3088-3096.

[5] R. Prabhavalkar, R. Alvarez, C. Parada, P. Nakkiran, and T. N. Sainath, "Automatic gain control and multi-style training for robust small-footprint keyword spotting with deep neural networks," in *IEEE Proceedings of the International Conference on Acoustics, Speech and Signal Processing*, pp. 4704–4708.

[6] S. Panchapagesan, M. Sun, A. Khare, S. Matsoukas, A. Mandal, B. Hoffmeister, and S. Vitaladevuni, "Multi-task learning and weighted cross-entropy for dnn-based keyword spotting," in *Proceedings of Interspeech*, 2016, pp. 760-764.

[7] T. N. Sainath and C. Parada, "Convolutional neural networks for small-footprint keyword spotting," in *Proceedings of Interspeech*, 2015, pp. 1478-1482

[8] Y. Wang, P. Getreuer, T. Hughes, R. F. Lyon, and R. A. Saurous, "Trainable frontend for robust and far-field keyword spotting," arXiv preprint, arXiv:1607.05666, 2016.

[9] K. Hwang, M. Lee, and W. Sung, "Online keyword spotting with a character-level recurrent neural network," arXiv preprint arXiv:1512.08903, 2015.

[10] S. Fernandez, A. Graves, and J. Schmidhuber, "An application of recurrent neural networks to discriminative keyword spotting," in *Artificial Neural Networks*. Springer, 2007, pp. 220–229.

[11] C. Lengerich, and A. Hannun, "An end-to-end architecture for keyword spotting and voice activity detection," *arXiv preprint arXiv:1611.09405*, 2016.

[12] L. Deng, and J. C. Platt, "Ensemble deep learning for speech recognition," in *Proceedings of Interspeech*, 2014.

[13] T. N. Sainath, O. Vinyals, A. Senior A, and H. Sak, "Convolutional, long short-term memory, fully connected deep neural networks," in *IEEE Proceedings of the International Conference on Acoustics, Speech and Signal Processing*, 2015, pp. 4580-4584.

[14] D. Amodei et al. "Deep Speech 2: End-to-end speech recognition in English and Mandarin.," *arXiv preprint arXiv:1512.02595*, 2015.

[15] K. Cho, B. van Merrienboer, D. Bahdanau, and Y. Bengio, "On the properties of neural machine translation: Encoder-decoder approaches," *arXiv preprint arXiv:1409.1259*, 2014.

[16] S. Hochreiter and J. Schmidhuber, "Long Short-Term Memory," *Neural Computation*, vol. 9, no. 8, pp. 1735-1780, 1997.

[17] D. Kingma, and J. Ba, "Adam: A method for stochastic optimization," *arXiv preprint arXiv:1412.6980*, 2014.

[18] J. Shelton, and G. P. Kumar GP, "Comparison between auditory and visual simple reaction times," *Neuroscience and medicine*, vol. 1 no. 1, pp. 30-32, 2010.

[19] K. Kumatani et al. "Microphone array processing for distant speech recognition: Towards real-world deployment," in *IEEE Asia-Pacific Signal & Information Processing Association Annual Summit and Conference*, 2012.